\documentclass[11pt,letterpaper]{article}
\usepackage{emnlp2017}
\usepackage{times}
\usepackage{latexsym}
\usepackage{amsfonts}
\usepackage{caption}
\usepackage{subcaption}
\usepackage{epsfig}
\usepackage{graphicx}
\usepackage{amssymb}
\usepackage{amsmath}
\usepackage{color}

\emnlpfinalcopy

\def\emnlppaperid{1151}

\newcommand{\cev}[1]{\reflectbox{\ensuremath{\vec{\reflectbox{\ensuremath{#1}}}}}}

\title{Hierarchically-Attentive RNN for Album Summarization and Storytelling}

\author{Licheng Yu \and Mohit Bansal \and Tamara L. Berg \\
UNC Chapel Hill\\
  {\tt \{licheng, mbansal, tlberg\}@cs.unc.edu}}

\date{}

\begin{document}

\maketitle

\begin{abstract}
We address the problem of end-to-end visual storytelling. 
Given a photo album, our model first selects the most representative (summary) photos, and then composes a natural language story for the album.
For this task, we make use of the Visual Storytelling dataset and a model composed of three hierarchically-attentive Recurrent Neural Nets (RNNs) to: encode the album photos, select representative (summary) photos, and  compose the story.
Automatic and human evaluations show our model achieves better performance on selection, generation, and retrieval than baselines. 
\end{abstract}

\vspace{-.2cm}
\section{Introduction}\label{sec_intro}
\vspace{-.2cm}

Since we first developed language, humans have always told stories. Fashioning a good story is an act of creativity and developing algorithms to replicate this has been a long running challenge. Adding pictures as input 
can provide information for guiding story construction by offering visual illustrations of the storyline.
In the related task of image captioning, most methods try to generate descriptions only for individual images or for short videos depicting a single activity.  Very recently, datasets have been introduced that extend this task to longer temporal sequences such as movies or photo albums~\cite{rohrbach2016movie,pan2016jointly,lu2013story,huang2016visual}. 

The type of data we consider in this paper provides input illustrations for story generation in the form of photo albums, sampled over a few minutes to a few days of time. For this type of data, generating textual descriptions involves telling a temporally consistent story about the depicted visual information, where stories must be coherent and take into account the temporal context of the images. Applications of this include constructing visual and textual summaries of albums, or even enabling search through personal photo collections to find photos of life events. 

Previous visual storytelling works can be classified into two types, vision-based and language-based, where image or language stories are constructed respectively.
Among the vision-based approaches, unsupervised learning is commonly applied:
e.g.,~\cite{sigurdsson2016learning} learns the latent temporal dynamics given a large amount of albums, and~\cite{kim2014reconstructing} formulate the photo selection as a sparse time-varying directed graph. 
However, these visual summaries tend to be difficult to evaluate and selected  photos may not agree with human selections. For language-based approaches, a sequence of natural language sentences are generated to describe a set of photos.
To drive this work ~\cite{park2015expressing} collected a dataset mined from Blog Posts. However, this kind of data often contains contextual information or loosely related language. A more direct dataset was recently released~\cite{huang2016visual}, where multi-sentence stories are collected describing photo albums via Amazon Mechanical Turk.

In this paper, we make use of the Visual Storytelling Dataset~\cite{huang2016visual}.
While the authors provide a seq2seq baseline, they only deal with the task of generating stories given 5-representative (summary) photos hand-selected by people from an album.
Instead, we focus on the more challenging and realistic problem of end-to-end generation of stories from entire albums. This requires us to either generate a story from all of the album's photos or to learn selection mechanisms to identify representative photos and then generate stories from those summary photos. We evaluate each type of approach. 

Ultimately, we propose a model of hierarchically-attentive recurrent neural nets, consisting of three RNN stages. The first RNN encodes the whole album context and each photo's content, the second RNN provides weights for photo selection, and the third RNN takes the weighted representation and decodes to the resulting sentences. Note that during training, we are only given the full input albums and the output stories, and our model needs to learn the summary photo selections latently.

We show that our model achieves better performance over baselines under both automatic  metrics and human evaluations. As a side product, we show that the latent photo selection also reasonably mimics human selections. Additionally, we propose an album retrieval task that can reliably pick the correct photo album given a sequence of sentences, and find that our model also outperforms the baselines on this task.

\begin{figure*}[t]
\centering
\vspace{-12pt}
\includegraphics[width=0.95\textwidth]{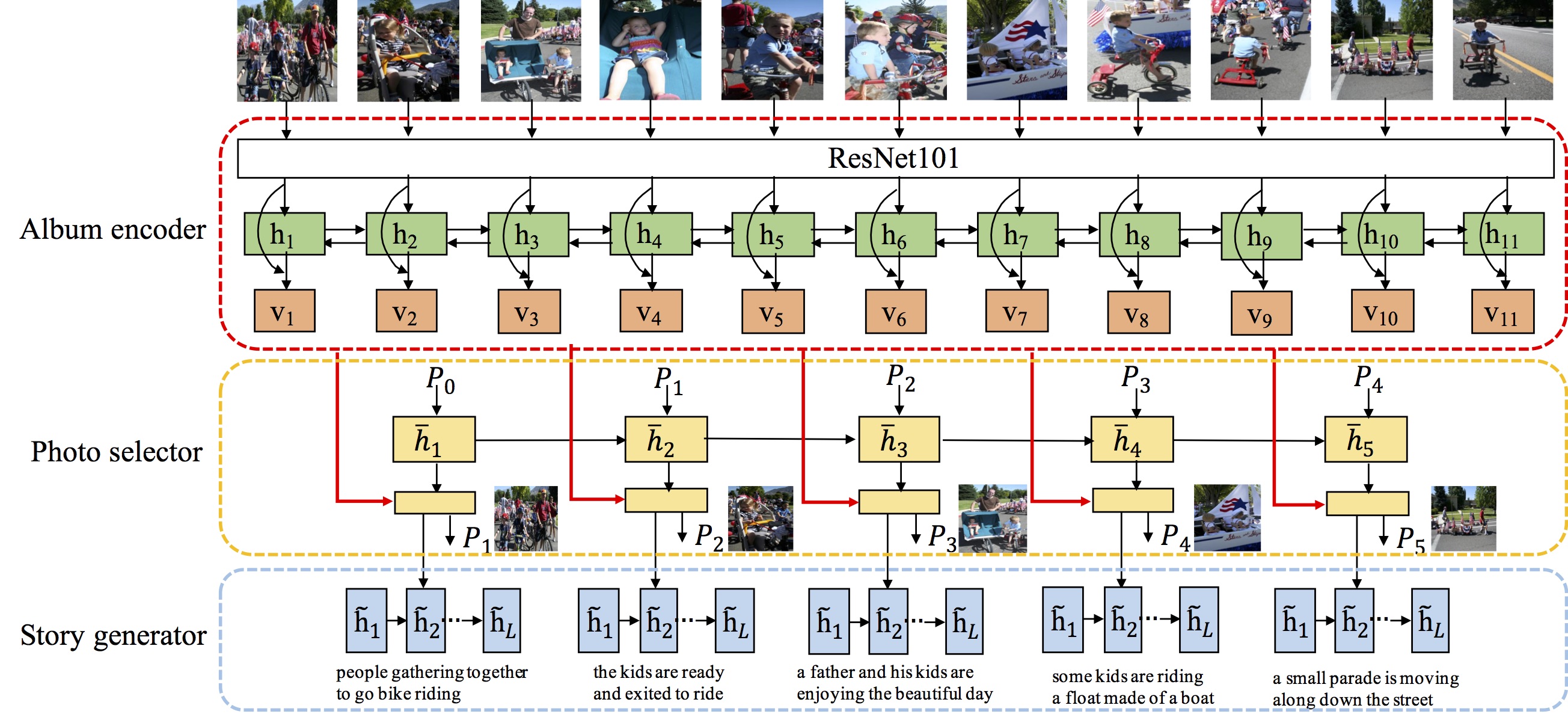}
\vspace{-0.3cm}
\caption{Model: the {\em album encoder} is a bi-directional GRU-RNN that encodes all album photos; the {\em photo selector} computes the probability of each photo being the $t$th album-summary photo; and finally, the {\em story generator} outputs a sequence of sentences that combine to tell a story for the album. 
}
\vspace{-.2cm}
\label{fig:model}
\end{figure*}

\vspace{-.2cm}
\section{Related work}
\vspace{-.2cm}

Recent years have witnessed an explosion of interest in vision and language tasks, reviewed below.

\noindent{\bf Visual Captioning:}
Most recent approaches to image captioning~\cite{vinyals2015show, xu2015show} have used CNN-LSTM structures to generate descriptions. For captioning video or movie content~\cite{venugopalan2015sequence, pan2016jointly}, sequence-to-sequence models are widely applied, where the first sequence encodes video frames and the second sequence decodes the description. Attention techniques~\cite{xu2015show, yu2016video, yao2015describing} are commonly incorporated for both tasks to localize salient temporal or spatial information.

\noindent{\bf Video Summarization:}
Similar to documentation summarization~\cite{rush2015neural, cheng2016neural, mei2016selective, woodsend2010automatic} which extracts key sentences and words, video summarization selects key frames or shots.
While some approaches use unsupervised learning~\cite{lu2013story, khosla2013large} or intuitive criteria to pick salient frames, recent models learn from human-created summaries~\cite{gygli2015video, zhang2016video, zhang2016summary, gong2014diverse}. 
Recently, to better exploit semantics,~\cite{choi2017textually} proposed textually customized summaries. 

\noindent{\bf Visual Storytelling:}
Visual storytelling tries to tell a coherent visual or textual story about an image set. 
Previous works include storyline graph modeling~\cite{kim2014reconstructing}, unsupervised mining~\cite{sigurdsson2016learning}, blog-photo alignment~\cite{kim2015joint}, 
and language re-telling~\cite{huang2016visual, park2015expressing}.
While~\cite{park2015expressing} collects data by mining Blog Posts,~\cite{huang2016visual} collects stories using Mechanical Turk, providing more directly relevant stories.

\vspace{-.2cm}
\section{Model}
\vspace{-.2cm}

Our model (Fig.~\ref{fig:model}) is composed of three modules: Album Encoder, Photo Selector, and Story Generator, jointly learned during training.

\subsection{Album Encoder}\label{sec:album_encoder}
\vspace{-.1cm}

Given an album $A=\{a_1, a_2, ..., a_n\}$, composed of a set of photos, we use a bi-directional RNN to encode the local album context for each photo.
We first extract the 2048-dimensional visual representation $f_i\in R^k$ for each photo using ResNet101~\cite{he2016deep}, then
a bi-directional RNN is applied to encode the full album.
Following~\cite{huang2016visual}, we choose a Gated Recurrent Unit (GRU) as the RNN unit to encode the photo sequence.
The sequence output at each time step encodes the local album context for each photo (from both directions).
Fused with the visual representation followed by ReLU, our final photo representation is (top module in Fig.~\ref{fig:model}):
\vspace{-.2cm}
\begin{equation}
\begin{split}\nonumber
f_i &= \mbox{ResNet}(a_i) \\ 
\vec{h}_i &= \vec{\mbox{GRU}}_{album}(f_i, \vec{h}_{i-1}) \\
\cev{h}_i &= \cev{\mbox{GRU}}_{album}(f_i, \cev{h}_{i+1}) \\
v_i &= \mbox{ReLU}([\vec{h}_i, \cev{h}_i ] + f_i).
\end{split}
\end{equation}
\vspace{-.3cm}

\vspace{-.4cm}
\subsection{Photo Selector}\label{sec:photo_selector}
\vspace{-.1cm}
The Photo Selector (illustrated in the middle yellow part of Fig.~\ref{fig:model}) identifies representative photos to summarize an album's content. As discussed, we do not assume that we are given the ground-truth album summaries during training, instead regarding selection as a latent variable in the end-to-end learning.
Inspired by Pointer Networks~\cite{vinyals2015pointer}, we use another GRU-RNN to perform this task \footnote{While the pointer network requires grounding labels, we regard the labels as latent variables}.

Given the album representation $V^{n\times k}$, the photo selector outputs probabilities $p_t\in R^n$ (likelihood of selection as $t$-th summary image) for all photos using soft attention. 
\vspace{-.1cm}
\begin{equation}\nonumber
\begin{split}
&\bar{h}_t = \mbox{GRU}_{select}(p_{t-1}, \bar{h}_{t-1}), \\
&p(y_{a_i}(t)=1) = \sigma(\mbox{MLP}([\bar{h}_t, v_i])),
\end{split}
\end{equation}
At each summarization step, $t$, the GRU takes the previous $p_{t-1}$ and previous hidden state as input, and outputs the next hidden state $\bar{h}_t$.
$\bar{h}_t$ is fused with each photo representation $v_i$ to compute the $i^{th}$ photo's attention $p_t^i = p(y_{a_i}(t)=1)$.
At test time, we simply pick the photo with the highest probability to be the summary photo at step $t$.

\subsection{Story Generator}
\vspace{-.1cm}

To generate an album's story, given the album representation matrix $V$ and photo summary probabilities $p_t$ from the first two modules, we compute the visual summary representation $g_t \in R^k$ (for the $t$-th summary step).
This is a weighted sum of the album representations, i.e., $g_t = p_t^T V$.  Each of these 5 $g_t$ embeddings (for $t = 1$ to $5$) is then used to decode 1 of the 5 story sentences respectively, as shown in the blue part of Fig.~\ref{fig:model}.

Given a story $S = \{s_t\}$, where $s_t$ is $t$-th summary sentence. 
Following~\newcite{donahue2015long}, the $l$-th word probability of the $t$-th sentence is:
\vspace{-.1cm}
\begin{equation}
\begin{split}
w_{t,l-1} &= W_e s_{t,l-1}, \\
\tilde{h}_{t, l} &= \mbox{GRU}_{story}(w_{t, l-1}, g_t, \tilde{h}_{t, l-1}),\\
p(s_{t,l}) &= \mbox{softmax}(\mbox{MLP}(\tilde{h}_{t, l})),
\end{split}
\end{equation}
\vspace{-.1cm}
where $W_e$ is the word embedding.
The GRU takes the joint input of visual summarization $g_t$, the previous word embedding $w_{t,l}$, and the previous hidden state, then outputs the next hidden state.
The generation loss is then the sum of the negative log likelihoods of the correct words: $L_{gen}(S) = -\sum_{t=1}^T\sum_{l=1}^{L_t} \log p_{t,l}(s_{t,l})$. 
\vspace{.1cm}

To further exploit the notion of temporal coherence in a story, we add an order-preserving constraint  to order the sequence of sentences within a story (related to the story-sorting idea in~\newcite{agrawal2016sort}).
For each story $S$ we randomly shuffle its 5 sentences to generate negative story instances $S'$.
We then apply a max-margin ranking loss to encourage correctly-ordered stories: $L_{rank}(S, S') = \max(0, m-\log p(S')+\log p(S))$.
The final loss is then a combination of the generation and ranking losses:  
\begin{equation}
L = L_{gen}(S) + \lambda L_{rank}(S, S').\label{eqn:loss}
\end{equation}

\section{Experiments}

We use the Visual Storytelling Dataset~\cite{huang2016visual}, consisting of 10,000 albums with 200,000 photos.
Each album contains 10-50 photos taken within a 48-hour span with two annotations: 1) 2 album summarizations, each with 5 selected representative photos, and 2) 5 stories describing the selected photos. 

\subsection{Story Generation}

This task is to generate a 5-sentence story describing an album.
We compare our model with two sequence-to-sequence baselines: 1) an encoder-decoder model (enc-dec), where the sequence of album photos is encoded and the last hidden state is fed into the decoder for story generation,
2) an encoder-attention-decoder model~\cite{xu2015show} (enc-attn-dec) with weights computed using a soft-attention mechanism.
At each decoding time step, a weighted sum of hidden states from the encoder is decoded. For fair comparison, we use the same album representation (Sec.~\ref{sec:album_encoder}) for the baselines.

\begin{table}[t]
\footnotesize
\begin{center}
\begin{tabular}{l|c c c c}
\multicolumn{5}{c}{beam size=3}\\
\hline
& Bleu3 & Rouge & Meteor & CIDEr \\
\hline
enc-dec & 19.58 & 29.23 & 33.02 & 4.65 \\
enc-attn-dec & 19.73 & 28.94 & 32.98 & 4.96 \\
h-attn & 20.53 & 29.82 & 33.81 & 6.84 \\
h-attn-rank & \bf{20.78} & \bf{29.82} & \bf{33.94} & \bf{7.38} \\
\hline
h-(gd)attn-rank & 21.02 & 29.53 & 34.12 & 7.51\\
\hline
\end{tabular}
\end{center}
\vspace{-.4cm}
\caption{Story generation evaluation.}
\vspace{-.1cm}
\label{table:generation}
\end{table}

\begin{table}[t]
\footnotesize
\begin{center}
\begin{tabular}{l | l}
\hline
enc-dec (29.50\%) & h-attn-rank (70.50\%)\\
\hline
enc-attn-dec (30.75\%) & h-attn-rank (69.25\%) \\
\hline
\hline
h-attn-rank (30.50\%) & gd-truth (69.50\%) \\ 
\hline
\end{tabular}
\end{center}
\vspace{-.4cm}
\caption{Human evaluation showing how often people prefer one model over the other.}
\vspace{-.2cm}
\label{table:human}
\end{table}

\begin{table}[t]
\footnotesize
\begin{center}
\begin{tabular}{l | c c }
\hline
& precision & recall \\
\hline
DPP & 43.75\% & 27.41\% \\
\hline
enc-attn-dec & 38.53\% & 24.25\% \\
\hline
h-attn 		 & 42.85\% & 27.10\% \\
\hline
h-attn-rank  & \bf{45.51}\% & \bf{28.77}\% \\
\hline
\end{tabular}
\end{center}
\vspace{-.4cm}
\caption{Album summarization evaluation.}
\vspace{-.1cm}
\label{table:summarization}
\end{table}

\begin{table}[t]
\footnotesize
\begin{center}
\begin{tabular}{l | c c c c}
\hline
& R@1 & R@5 & R@10 & MedR \\
\hline
enc-dec      & 10.70\% & 29.30\% & 41.40\% & 14.5 \\
\hline
enc-attn-dec & 11.60\% & 33.00\% & 45.50\% & 11.0 \\
\hline  
h-attn       & 18.30\% & \bf{44.50}\% & \bf{57.60}\% & \bf{6.0} \\
\hline
h-attn-rank  & \bf{18.40}\% & 43.30\% & 55.50\% & 7.0 \\
\hline
\end{tabular}
\end{center}
\vspace{-.4cm}
\caption{1000 album retrieval evaluation.}
\vspace{-.1cm}
\label{table:retrieval}
\end{table}

We test two variants of our model trained with and without ranking regularization by controlling $\lambda$ in our loss function, denoted as h-attn (without ranking), and h-attn-rank (with ranking).
Evaluations of each model are shown in Table~\ref{table:generation}.
The h-attn outperforms both baselines, and h-attn-rank achieves the best performance for all metrics.
Note, we use beam-search with beam size=3 during generation for a reasonable performance-speed trade-off (we observe similar improvement trends with beam size = 1).\footnote{We also compute the $p$-value of Meteor on 100K samples via the bootstrap test~\cite{efron1994introduction}, as Meteor has better agreement with human judgments than Bleu/Rouge~\cite{huang2016visual}.
Our h-attn-rank model has strong statistical significance ($p=0.01$) over the enc-dec and enc-attn-dec models (and is similar to the h-attn model).}
To test performance under optimal image selection, we use one of the two ground-truth human-selected 5-photo-sets as an oracle to hard-code the photo selection, denoted as h-(gd)attn-rank.
This achieves only a slightly higher Meteor compared to our end-to-end model.

Additionally, we also run human evaluations in a forced-choice task where people choose between stories generated by different methods.
For this evaluation, we select 400 albums, each evaluated by 3 Turkers. Results are shown in Table~\ref{table:human}. 
Experiments find significant preference for our model over both baselines. As a simple Turing test, we also compare our results with human written stories (last row of Table~\ref{table:human}), indicating room for improvement of methods.

\subsection{Album Summarization}
We evaluate the precision and recall of our generated summaries (output by the photo selector) compared to human selections (the combined set of both human-selected 5-photo stories). 
For comparison, we evaluate enc-attn-dec on the same task by aggregating predicted attention and selecting the 5 photos with highest accumulated attention.
Additionally, we also run DPP-based video summarization~\cite{kulesza2012determinantal} using the same album features.
Our models have higher performance compared to baselines as shown in Table~\ref{table:summarization} (though DPP also achieves strong results, indicating that there is still room to improve the pointer network).

\vspace{-0.2cm}
\subsection{Output Example Analysis}
\begin{figure*}[t!]
\centering
\includegraphics[width=0.98\textwidth]{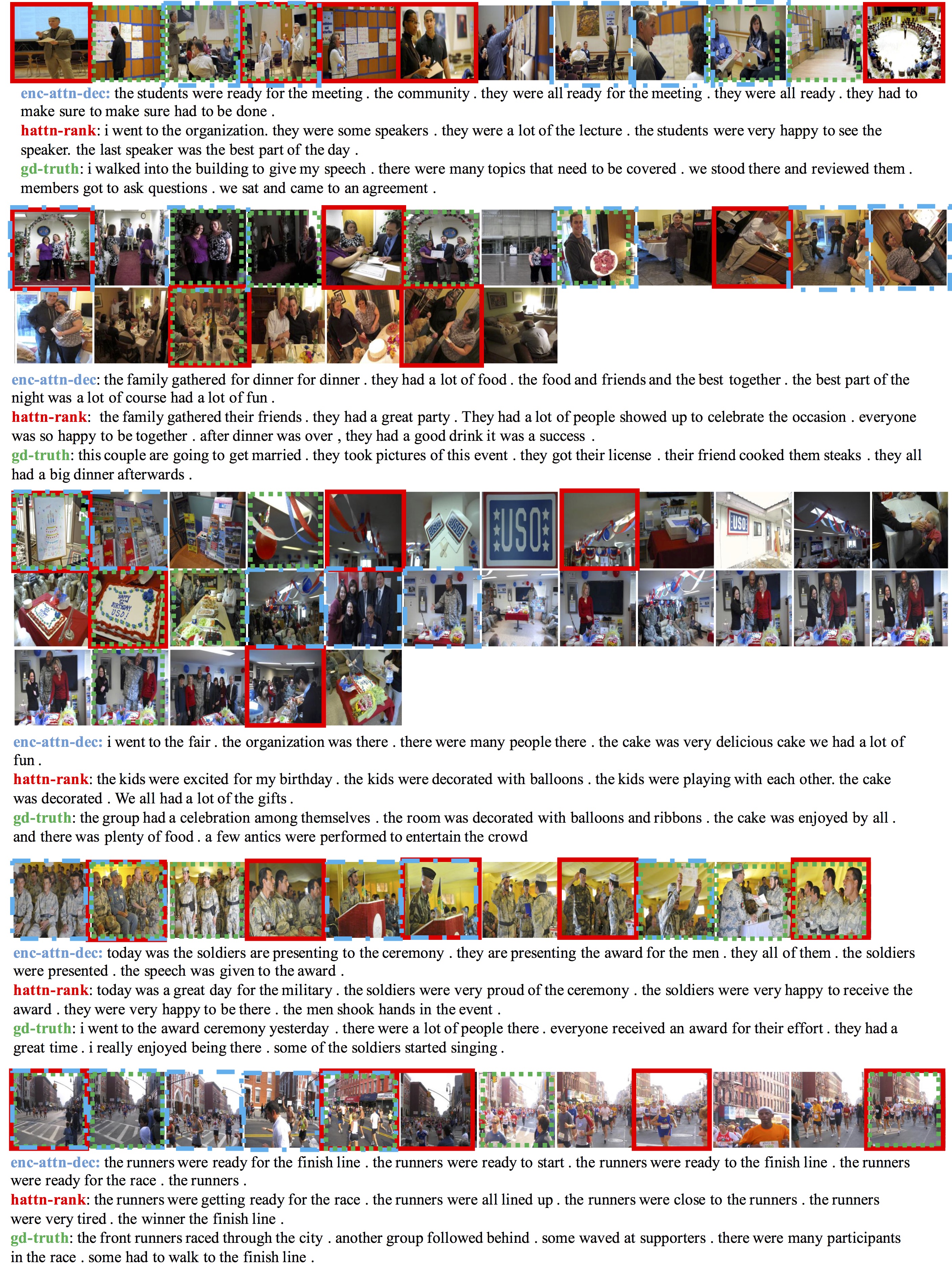}
\caption{Examples of album summarization and storytelling by enc-attn-dec (blue), h-attn-rank (red), and ground-truth (green). We randomly select 1 out of 2 human album summaries as ground-truth here.}
\label{fig:generation_a}
\end{figure*}

\begin{figure*}[t!]
\centering
\includegraphics[width=0.98\textwidth]{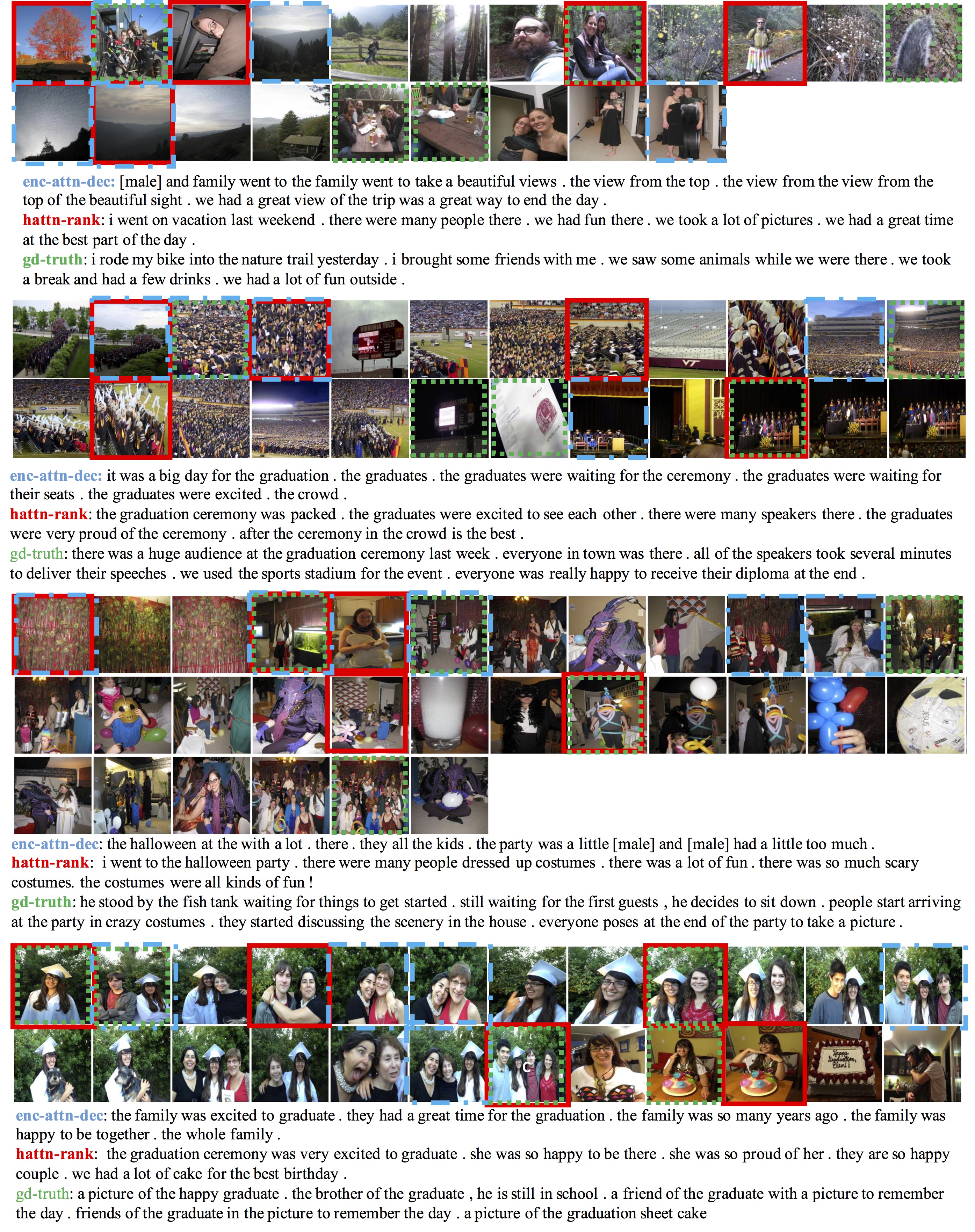}
\caption{More examples of album summarization and storytelling by enc-attn-dec (blue), h-attn-rank (red), and ground-truth (green). We randomly select 1 out of 2 human album summaries as ground-truth here.}
\label{fig:generation_b}
\end{figure*}

Fig.~\ref{fig:generation_a} and Fig.~\ref{fig:generation_b} shows several output examples of the joint album summarization and storytelling generation.
We compare our full model h-attn-rank with the baseline enc-attn-dec, as both models are able to do the album summarization and story generation tasks jointly.
In Fig.~\ref{fig:generation_a} and Fig.~\ref{fig:generation_b}, we use blue dashed box and red box to indicate the album summarization by the two models respectively.
As reference, we also show the ground-truth album summaries by randomly selecting 1 out of 2 human album summaries, which are highlighted with green box.
Below each album are their generated stories.

\subsection{Album Retrieval}
Given a human-written story, we introduce a task to retrieve the album described by that story. 
We randomly select 1000 albums and one ground-truth story from each for evaluation. 
Using the generation loss, we compute the likelihood of each album $A_m$ given the query story $S$ and retrieve the album with the highest generation likelihood, $A = \mbox{argmax}_{A_m} p(S|A_m)$.
We use Recall@k and Median Rank for evaluation.
As shown in Table~\ref{table:retrieval}), we find that our models outperform the baselines, but the ranking term in Eqn.\ref{eqn:loss} does not improve performance significantly.

\section{Conclusion}
\vspace{-0.3cm}
Our proposed hierarchically-attentive RNN based models for end-to-end visual storytelling can jointly summarize and generate relevant stories from full input photo albums effectively. Automatic and human evaluations show that our method outperforms strong sequence-to-sequence baselines on selection, generation, and retrieval tasks.

\vspace{0cm}
\section*{Acknowledgments}
\vspace{-0.4cm}
We thank the anonymous reviewers for their helpful comments. This research is supported by NSF Awards \#1633295, 1444234, 1445409, 1562098.

\bibliography{emnlp2017}

\begin{thebibliography}{4}
\expandafter\ifx\csname natexlab\endcsname\relax\def\natexlab#1{#1}\fi

\bibitem[{Aho and Ullman(1972)}]{Aho:72}
Alfred~V. Aho and Jeffrey~D. Ullman. 1972.
\newblock \emph{The Theory of Parsing, Translation and Compiling}, volume~1.
\newblock Prentice-Hall, Englewood Cliffs, NJ.

\bibitem[{{American Psychological Association}(1983)}]{APA:83}
{American Psychological Association}. 1983.
\newblock \emph{Publications Manual}.
\newblock American Psychological Association, Washington, DC.

\bibitem[{Chandra et~al.(1981)Chandra, Kozen, and Stockmeyer}]{Chandra:81}
Ashok~K. Chandra, Dexter~C. Kozen, and Larry~J. Stockmeyer. 1981.
\newblock \href {https://doi.org/10.1145/322234.322243} {Alternation}.
\newblock \emph{Journal of the Association for Computing Machinery},
  28(1):114--133.

\bibitem[{Gusfield(1997)}]{Gusfield:97}
Dan Gusfield. 1997.
\newblock \emph{Algorithms on Strings, Trees and Sequences}.
\newblock Cambridge University Press, Cambridge, UK.

\end{thebibliography}


\begin{thebibliography}{29}
\expandafter\ifx\csname natexlab\endcsname\relax\def\natexlab#1{#1}\fi

\bibitem[{Agrawal et~al.(2016)Agrawal, Chandrasekaran, Batra, Parikh, and
  Bansal}]{agrawal2016sort}
Harsh Agrawal, Arjun Chandrasekaran, Dhruv Batra, Devi Parikh, and Mohit
  Bansal. 2016.
\newblock Sort story: Sorting jumbled images and captions into stories.
\newblock In \emph{EMNLP}.

\bibitem[{Cheng and Lapata(2016)}]{cheng2016neural}
Jianpeng Cheng and Mirella Lapata. 2016.
\newblock Neural summarization by extracting sentences and words.
\newblock In \emph{ACL}.

\bibitem[{Choi et~al.(2017)Choi, Oh, and Kweon}]{choi2017textually}
Jinsoo Choi, Tae-Hyun Oh, and In~So Kweon. 2017.
\newblock Textually customized video summaries.
\newblock \emph{arXiv preprint arXiv:1702.01528}.

\bibitem[{Donahue et~al.(2015)Donahue, Anne~Hendricks, Guadarrama, Rohrbach,
  Venugopalan, Saenko, and Darrell}]{donahue2015long}
Jeffrey Donahue, Lisa Anne~Hendricks, Sergio Guadarrama, Marcus Rohrbach,
  Subhashini Venugopalan, Kate Saenko, and Trevor Darrell. 2015.
\newblock Long-term recurrent convolutional networks for visual recognition and
  description.
\newblock In \emph{CVPR}.

\bibitem[{Efron and Tibshirani(1994)}]{efron1994introduction}
Bradley Efron and Robert~J Tibshirani. 1994.
\newblock \emph{An introduction to the bootstrap}.
\newblock CRC press.

\bibitem[{Gong et~al.(2014)Gong, Chao, Grauman, and Sha}]{gong2014diverse}
Boqing Gong, Wei-Lun Chao, Kristen Grauman, and Fei Sha. 2014.
\newblock Diverse sequential subset selection for supervised video
  summarization.
\newblock In \emph{NIPS}.

\bibitem[{Gygli et~al.(2015)Gygli, Grabner, and Van~Gool}]{gygli2015video}
Michael Gygli, Helmut Grabner, and Luc Van~Gool. 2015.
\newblock Video summarization by learning submodular mixtures of objectives.
\newblock In \emph{CVPR}.

\bibitem[{He et~al.(2016)He, Zhang, Ren, and Sun}]{he2016deep}
Kaiming He, Xiangyu Zhang, Shaoqing Ren, and Jian Sun. 2016.
\newblock Deep residual learning for image recognition.
\newblock In \emph{CVPR}.

\bibitem[{Huang et~al.(2016)Huang, Ferraro, Mostafazadeh, Misra, Agrawal,
  Devlin, Girshick, He, Kohli, Batra et~al.}]{huang2016visual}
Ting-Hao~Kenneth Huang, Francis Ferraro, Nasrin Mostafazadeh, Ishan Misra,
  Aishwarya Agrawal, Jacob Devlin, Ross Girshick, Xiaodong He, Pushmeet Kohli,
  Dhruv Batra, et~al. 2016.
\newblock Visual storytelling.
\newblock In \emph{NACCL}.

\bibitem[{Khosla et~al.(2013)Khosla, Hamid, Lin, and
  Sundaresan}]{khosla2013large}
Aditya Khosla, Raffay Hamid, Chih-Jen Lin, and Neel Sundaresan. 2013.
\newblock Large-scale video summarization using web-image priors.
\newblock In \emph{CVPR}.

\bibitem[{Kim et~al.(2015)Kim, Moon, and Sigal}]{kim2015joint}
Gunhee Kim, Seungwhan Moon, and Leonid Sigal. 2015.
\newblock Joint photo stream and blog post summarization and exploration.
\newblock In \emph{CVPR}.

\bibitem[{Kim and Xing(2014)}]{kim2014reconstructing}
Gunhee Kim and Eric~P Xing. 2014.
\newblock Reconstructing storyline graphs for image recommendation from web
  community photos.
\newblock In \emph{CVPR}.

\bibitem[{Kulesza et~al.(2012)Kulesza, Taskar
  et~al.}]{kulesza2012determinantal}
Alex Kulesza, Ben Taskar, et~al. 2012.
\newblock Determinantal point processes for machine learning.
\newblock \emph{Foundations and Trends{\textregistered} in Machine Learning}.

\bibitem[{Lu and Grauman(2013)}]{lu2013story}
Zheng Lu and Kristen Grauman. 2013.
\newblock Story-driven summarization for egocentric video.
\newblock In \emph{CVPR}.

\bibitem[{Mei et~al.(2016)Mei, Bansal, and Walter}]{mei2016selective}
Hongyuan Mei, Mohit Bansal, and Matthew~R. Walter. 2016.
\newblock What to talk about and how? selective generation using lstms with
  coarse-to-fine alignment.
\newblock In \emph{NAACL}.

\bibitem[{Pan et~al.(2016)Pan, Mei, Yao, Li, and Rui}]{pan2016jointly}
Yingwei Pan, Tao Mei, Ting Yao, Houqiang Li, and Yong Rui. 2016.
\newblock Jointly modeling embedding and translation to bridge video and
  language.
\newblock In \emph{CVPR}.

\bibitem[{Park and Kim(2015)}]{park2015expressing}
Cesc~C Park and Gunhee Kim. 2015.
\newblock Expressing an image stream with a sequence of natural sentences.
\newblock In \emph{NIPS}.

\bibitem[{Rohrbach et~al.(2016)Rohrbach, Torabi, Rohrbach, Tandon, Pal,
  Larochelle, Courville, and Schiele}]{rohrbach2016movie}
Anna Rohrbach, Atousa Torabi, Marcus Rohrbach, Niket Tandon, Christopher Pal,
  Hugo Larochelle, Aaron Courville, and Bernt Schiele. 2016.
\newblock Movie description.
\newblock \emph{IJCV}.

\bibitem[{Rush et~al.(2015)Rush, Chopra, and Weston}]{rush2015neural}
Alexander~M Rush, Sumit Chopra, and Jason Weston. 2015.
\newblock A neural attention model for abstractive sentence summarization.
\newblock In \emph{EMNLP}.

\bibitem[{Sigurdsson et~al.(2016)Sigurdsson, Chen, and
  Gupta}]{sigurdsson2016learning}
Gunnar~A Sigurdsson, Xinlei Chen, and Abhinav Gupta. 2016.
\newblock Learning visual storylines with skipping recurrent neural networks.
\newblock In \emph{ECCV}.

\bibitem[{Venugopalan et~al.(2015)Venugopalan, Rohrbach, Donahue, Mooney,
  Darrell, and Saenko}]{venugopalan2015sequence}
Subhashini Venugopalan, Marcus Rohrbach, Jeffrey Donahue, Raymond Mooney,
  Trevor Darrell, and Kate Saenko. 2015.
\newblock Sequence to sequence-video to text.
\newblock In \emph{ICCV}.

\bibitem[{Vinyals et~al.(2015{\natexlab{a}})Vinyals, Fortunato, and
  Jaitly}]{vinyals2015pointer}
Oriol Vinyals, Meire Fortunato, and Navdeep Jaitly. 2015{\natexlab{a}}.
\newblock Pointer networks.
\newblock In \emph{NIPS}.

\bibitem[{Vinyals et~al.(2015{\natexlab{b}})Vinyals, Toshev, Bengio, and
  Erhan}]{vinyals2015show}
Oriol Vinyals, Alexander Toshev, Samy Bengio, and Dumitru Erhan.
  2015{\natexlab{b}}.
\newblock Show and tell: A neural image caption generator.
\newblock In \emph{CVPR}.

\bibitem[{Woodsend and Lapata(2010)}]{woodsend2010automatic}
Kristian Woodsend and Mirella Lapata. 2010.
\newblock Automatic generation of story highlights.
\newblock In \emph{ACL}.

\bibitem[{Xu et~al.(2015)Xu, Ba, Kiros, Cho, Courville, Salakhutdinov, Zemel,
  and Bengio}]{xu2015show}
Kelvin Xu, Jimmy Ba, Ryan Kiros, Kyunghyun Cho, Aaron~C Courville, Ruslan
  Salakhutdinov, Richard~S Zemel, and Yoshua Bengio. 2015.
\newblock Show, attend and tell: Neural image caption generation with visual
  attention.
\newblock In \emph{ICML}.

\bibitem[{Yao et~al.(2015)Yao, Torabi, Cho, Ballas, Pal, Larochelle, and
  Courville}]{yao2015describing}
Li~Yao, Atousa Torabi, Kyunghyun Cho, Nicolas Ballas, Christopher Pal, Hugo
  Larochelle, and Aaron Courville. 2015.
\newblock Describing videos by exploiting temporal structure.
\newblock In \emph{ICCV}.

\bibitem[{Yu et~al.(2016)Yu, Wang, Huang, Yang, and Xu}]{yu2016video}
Haonan Yu, Jiang Wang, Zhiheng Huang, Yi~Yang, and Wei Xu. 2016.
\newblock Video paragraph captioning using hierarchical recurrent neural
  networks.
\newblock In \emph{CVPR}.

\bibitem[{Zhang et~al.(2016{\natexlab{a}})Zhang, Chao, Sha, and
  Grauman}]{zhang2016summary}
Ke~Zhang, Wei-Lun Chao, Fei Sha, and Kristen Grauman. 2016{\natexlab{a}}.
\newblock Summary transfer: Exemplar-based subset selection for video
  summarization.
\newblock In \emph{CVPR}.

\bibitem[{Zhang et~al.(2016{\natexlab{b}})Zhang, Chao, Sha, and
  Grauman}]{zhang2016video}
Ke~Zhang, Wei-Lun Chao, Fei Sha, and Kristen Grauman. 2016{\natexlab{b}}.
\newblock Video summarization with long short-term memory.
\newblock In \emph{ECCV}.

\end{thebibliography}
\bibliographystyle{emnlp_natbib}

\end{document}